\def\year{2019}\relax
\documentclass[letterpaper]{article} % DO NOT CHANGE THIS
\usepackage{arxiv}  % DO NOT CHANGE THIS
\usepackage[hyphens]{url}  % DO NOT CHANGE THIS
\usepackage{graphicx} % DO NOT CHANGE THIS
\usepackage{graphicx}  % DO NOT CHANGE THIS
\usepackage{natbib}

\usepackage{url}  %Required
\usepackage[hang,flushmargin]{footmisc}
\usepackage{graphicx}  %Required
\usepackage{subcaption}
\usepackage{subcaption}
\usepackage{multirow}
\usepackage{makecell}

\usepackage{algorithm}
\usepackage{algpseudocode}

\usepackage{enumitem}
\usepackage{mathtools}

\usepackage{Definitions}
\usepackage{color}

\makeatletter
\def\thm@space@setup{%
  \thm@preskip=\parskip \thm@postskip=0pt
}
\makeatother

\usepackage{amsmath}
\usepackage[normalem]{ulem}
\usepackage{amssymb}
\usepackage{booktabs}
\usepackage{bbm}

\usepackage{stmaryrd}

\usepackage{nameref}

\usepackage{mathabx}

\DeclareMathOperator{\maxp}{\text{maxpool}}

\begin{document} 

\title{RecurJac: An Efficient Recursive Algorithm for Bounding Jacobian Matrix of Neural Networks and Its Applications}
\author{Huan Zhang\thanks{Work done during internship at Microsoft Research}\\
UCLA\\
404 Westwood Plaza\\
Los Angeles, CA 90095\\
%\texttt
{huan@huan-zhang.com}
\And
Pengchuan Zhang\\
Microsoft Research AI\\
14820 NE 36th St\\
Redmond, WA 98052\\
%\texttt
{penzhan@microsoft.com}
\And
Cho-Jui Hsieh\\
UCLA\\
404 Westwood Plaza\\
Los Angeles, CA 90095\\
%\texttt
{chohsieh@cs.ucla.edu}
}
\maketitle

\begin{abstract}  
The Jacobian matrix (or the gradient for single-output networks) is directly related to many important properties of neural networks, such as the function landscape, stationary points, (local) Lipschitz constants and robustness to adversarial attacks. In this paper, we propose a \textit{recursive} algorithm, \textbf{RecurJac}, to compute both upper and lower bounds for each element in the Jacobian matrix of a neural network with respect to network's input, and the network can contain a wide range of activation functions. As a byproduct, we can efficiently obtain a (local) Lipschitz constant, which plays a crucial role in neural network robustness verification, as well as the training stability of GANs. Experiments show that (local) Lipschitz constants produced by our method is of better quality than previous approaches, thus providing better robustness verification results. Our algorithm has polynomial time complexity, and its computation time is reasonable even for relatively large networks. Additionally, we use our bounds of Jacobian matrix to characterize the landscape of the neural network, for example, to determine whether there exist stationary points in a local neighborhood. Source code available at \url{http://github.com/huanzhang12/RecurJac-Jacobian-bounds}.
\end{abstract} 

\section{Introduction}

%In this paper, we propose an efficient algorithm for computing the upper and lower bounds of the Jacobian matrix of deep neural networks. 
Deep neural networks have been successfully applied to many applications, but one of
the major criticisms is their being black boxes---no satisfactory explanation of their behavior can be easily offered. Given a neural network $f(\cdot)$ with input $x$, one fundamental question to ask is: how does a perturbation in the input space affect the output prediction? To formally answer this question and bound the behavior of neural networks, a critical step is to compute the uniform bounds of the Jacobian matrix $\frac{\partial f(x)}{\partial x}$ for all $x$ within a certain region. 
Many recent works on understanding or verifying the behavior of neural networks rely on this quantity. 
For example, once a (local) Jacobian bound is computed, one can immediately know the radius of a guaranteed ``safe region'' in the input space, where no adversarial perturbation can change the output label~\citep{hein2017formal,weng2018evaluating}. This is also referred to as the {\it robustness verification problem}. In generative adversarial networks (GANs)~\citep{goodfellow2014generative}, the training process suffers from the gradient vanishing problem and can be very unstable. Adding the Lipschitz constant of the discriminator network as a constraint~\citep{arjovsky2017wasserstein,miyato2018spectral} or as a regularizer~\citep{gulrajani2017improved} significantly improves the training stability of GANs. For neural networks, the Jacobian matrix $\frac{\partial f(x)}{\partial x}$ is also closely related to its Jacobian matrix with respect to the weights $\frac{\partial f(x; W)}{\partial W}$, whose bound directly characterizes the generalization gap in supervised learning and GANs; see, e.g., \cite{vapnik1998statistical,sriperumbudur2009integral,bartlett2017spectrally,arora2017gans,zhang2017discrimination}.  

%Computing bounds for Jacobian (or gradient) is very challenging even for a simple ReLU network, and 
How to efficiently provide a tight bound for Jacobian (or gradient) is still an open problem for deep neural networks. 
%Previous  attempts for computing Jacobian bounds can be summarized into three categories. 
Sampling-based approaches~\citep{wood1996estimation,weng2018evaluating} cannot provide a certified bound and the computed quantity is usually an under-estimation; bounding the norm of Jacobian matrix over the entire domain (i.e. global Lipschitz constant) by the product of operator norms of the weight matrices~\citep{szegedy2013intriguing,cisse2017parseval,elsayed2018large} produces a very loose global upper bound, especially when we are only interested in a small local region of a neural network. Additionally, some recent works focus on computing Lipschitz constant in ReLU networks: \citet{raghunathan2018certified} solves a semi-definite programming (SDP) problem to give a Lipschitz constant, but its computational cost is high and it only applies to 2-layer networks; Fast-Lip by \citet{weng2018evaluating} can be applied to multi-layer ReLU networks but the bound quickly loses its power when the network goes deeper.

In this paper, we propose a novel {\it recursive} algorithm, dubbed {\it RecurJac}, for efficiently computing a certified {\it Jacobian} bound.
Unlike the layer-by-layer algorithm (Fast-Lip) for ReLU network in \citep{weng2018evaluating}, we develop a recursive refinement procedure that significantly outperforms Fast-Lip on ReLU networks, and our algorithm is general enough to be applied to networks with most common activation functions, not limited to ReLU. Our key observation is that the Jacobian bounds of previous layers can be used to reduce the uncertainties of neuron activations in the current layer, and some uncertain neurons can be fixed without affecting the final bound. We can then absorb these fixed neurons into the previous layers' weight matrix, which results in bounding Jacobian matrix for another shallower network. This technique can be applied recursively to get a tighter final bound. 
Compared with the non-recursive algorithm (Fast-Lip), RecurJac increases the computation cost by at most $H$ times ($H$ is depth of the network), which is reasonable even for relatively large networks. 

\begin{figure}[t]
  \centering
  \includegraphics[width=0.36\textwidth]{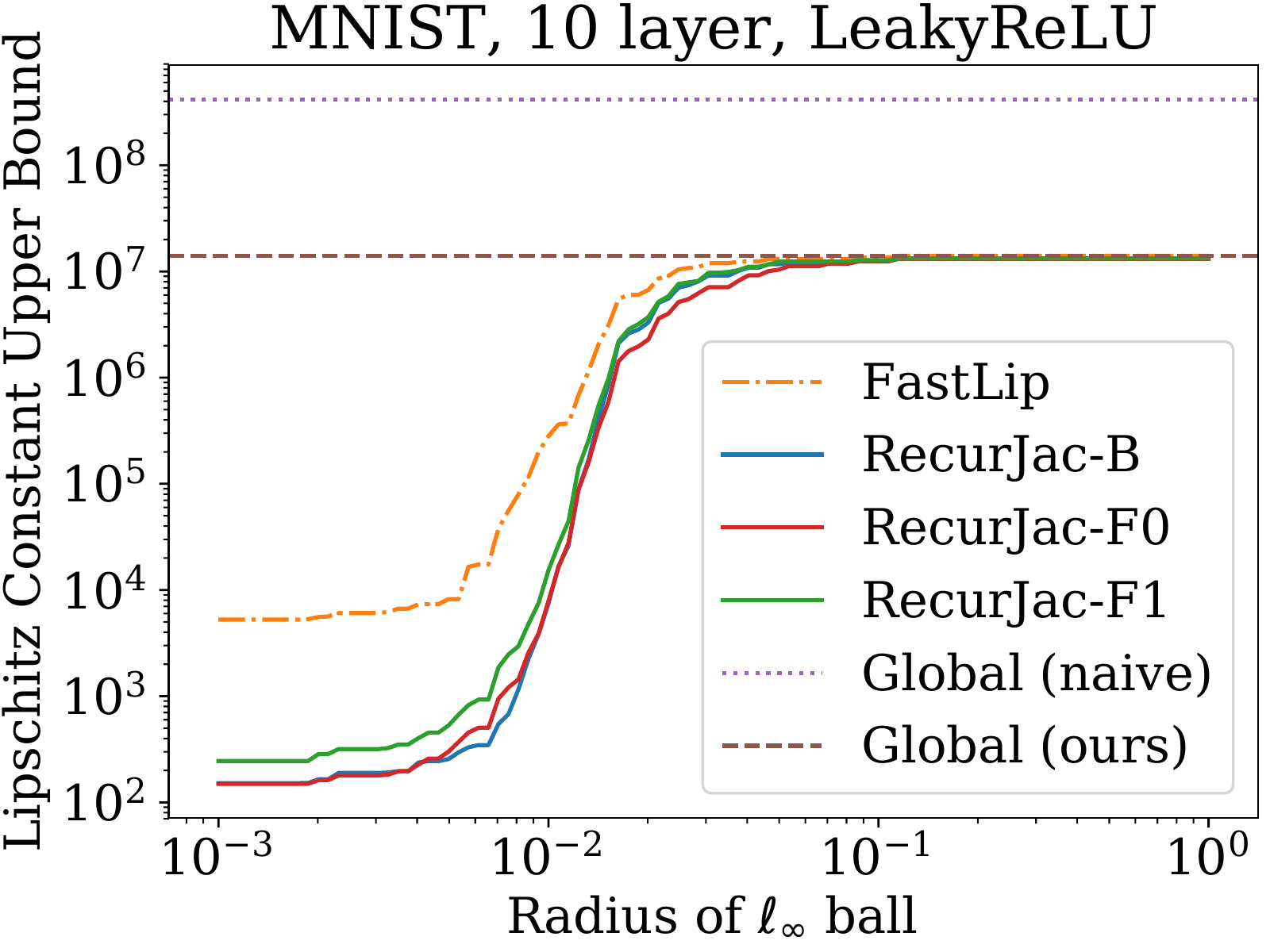}
  \caption{RecurJac can obtain local and global Lipschitz constants which are magnitudes better than existing algorithms. See Experiment section for more details.}
  \label{fig:lips_small}
\end{figure}

We apply RecurJac to various applications. First, we can investigate the local optimization landscape after obtaining the upper and lower bounds of Jacobian matrix, by guaranteeing that no stationary points exist inside a certain region. Experimental results show that the radius of this region steadily decreases when networks become deeper. Second, RecurJac can find a local Lipschitz constant, which up to two magnitudes smaller than the state-of-the-art algorithm without a recursive structure (Figure~\ref{fig:lips_small}). Finally, we can use RecurJac to evaluate the robustness of neural networks, by giving a certified lower bound within which no adversarial examples can be found.

\section{Related Work}

\paragraph{Previous algorithms for computing Lipschitz constant.}
Several previous works give bounds of the local or global Lipschitz constant of neural networks, which is a special case of our problem -- after knowing the element-wise bounds for Jacobian matrix, a local or global Lipschitz constant can be obtained by taking the corresponding induced norm of Jacobian matrix (more details in the next section).

One simple approach for estimating the Lipschitz constant for any black-box function is to sample many $x, y$ and
compute the maximal $\|f(x)-f(y)\|/\|x-y\|$~\citep{wood1996estimation}. However, the computed value may be an under-estimation unless the sample size goes to infinity.  The Extreme Value Theory~\citep{de2007extreme} can be used to refine the bound but the computed value could still under-estimate the Lipschitz constant~\citep{weng2018evaluating}, especially due to the high dimensionality of inputs.

For a neural network with known structure and weights, it is possible to compute Lipschitz constant explicitly. Since the Jacobian matrix of a neural network with respect to the input $x$ can be written explicitly (as we will introduce later in Eq.~\eqref{eq:gradient}), an easy and popular way to obtain a loose global Lipschitz constant is to multiply weight matrices' operator norms and the maximum derivative of each activation function (most common activation functions are Lipschitz continuous)~\citep{szegedy2013intriguing}. Since this quantity is simple to compute and can be optimized by back-propagation, many recent works also propose defenses to adversarial examples~\citep{cisse2017parseval,elsayed2018large,tsuzuku2018lipschitz,qian2018l2} or techniques to improve the training stability of GAN~\citep{miyato2018spectral} by regularizing this global Lipschitz constant.
%on the right hand side, assuming upper bounds for each $\Sigma$ can be computed.
However, it is clearly a very loose Lipschitz constant, as will be shown in our experiments.

For 2-layer ReLU networks, \citet{raghunathan2018certified} computes a global Lipschitz constant by relaxing the problem to semi-definite programming (SDP) and solving its dual, but it is computationally expensive. For 2-layer networks with twice differentiable activation functions, \citet{hein2017formal} derives the local Lipschitz constant for robustness verification. These methods show promising results for 2-layer networks, but cannot be trivially extended to networks with multiple layers.

\paragraph{Bounds for Jacobian matrix}
Recently, \citet{weng2018towards} proposes an layer-by-layer algorithm, Fast-Lip, for computing the lower and upper bounds of Jacobian matrix with respect to network input $x$. It exploits the special activation patterns in ReLU networks but does not apply to networks with general activation functions. Most importantly, it loses power quickly when the network becomes deeper. Using Fast-Lip for robustness verification produces non-trivial bounds only for very shallow networks (less than 4 layers).

\paragraph{Robustness verification of neural networks. }
Verifying if a neural network is robust to a norm bounded distortion is a NP-complete problem~\citep{katz2017reluplex}. Solving the minimal adversarial perturbation can take hours to days using solvers for satisfiability modulo theories (SMT)~\citep{katz2017reluplex} or mixed integer linear programming (MILP)~\citep{tjeng2017evaluating}. However, a lower bound for the minimum adversarial distortion (\textit{robustness lower bound}) can be given if knowing the local Lipschitz constant near the input example $x$.
For a multi-class classification network, assume the output of network 
$f(x)$ is a $K$-dimensional vector where each $f_j(x)$ is the logit for the $j$-th class and the final prediction $F(x)=\arg\max_j f_j(x)$, the following lemma gives a robustness lower bound~\citep{hein2017formal,weng2018evaluating}: 
\begin{lemma}
For an input example $x$, 
\begin{equation}
    F(x+\Delta) = y \ \ \text{for all} \ \|\Delta\|<\min\big\{ R, 
    \min_{j\neq y}\frac{f_y(x) - f_j(x)}{L_j}\big\}, 
    \label{eq:clever}
\end{equation}
where $L_j$ is the Lipschitz constant of $f_j(x)-f_y(x)$ in some local region (will be formally defined later).
\end{lemma}
Therefore, as long as a local Lipschitz constant can be computed, we can verify that the prediction of
a neural network will stay unchanged for any perturbation within radius $R$. 
A good local Lipschitz constant is hard to compute in general: \citet{hein2017formal} only show the results
for 2-layer neural networks; \citet{weng2018evaluating} apply a sampling-based approach and cannot guarantee
that the computed radius satisfies~\eqref{eq:clever}. Thus, an efficient, guaranteed and tight bound for Lipschitz constant is essential for understanding the robustness of deep neural networks.

Some other methods have also been proposed for robustness verification, including direct linear bounds~\citep{zhang2018crown,croce2018provable,weng2018towards}, convex adversarial polytope~\citep{wong2018provable,wong2018scaling},  Lagrangian dual relaxation~\citep{dvijotham2018dual} and geometry abstraction~\citep{gehr2018ai,mirman2018differentiable}. In this paper we focus on Local Lipschitz constant based methods only.

\section{RecurJac: Recursive Jacobian Bounding} 

% Define symbols
\newcommand{\tail}[2]{#1_{\overline{[#2]}}}
\newcommand{\abs}[1]{|#1|}
\newcommand{\tabs}[1]{\left|#1\right|}
\newcommand{\wh}{\widehat}
\newcommand{\wt}{\widetilde}
\newcommand{\ov}{\overline}
\newcommand{\eps}{\epsilon}
\newcommand{\N}{\mathcal{N}}
\newcommand{\R}{\mathbb{R}}
\newcommand{\volume}{\mathrm{volume}}
\newcommand{\RHS}{\mathrm{RHS}}
\newcommand{\LHS}{\mathrm{LHS}}
\newcommand{\bone}{\mathbf{1}}
\renewcommand{\i}{\mathbf{i}}
\newcommand{\norm}[1]{\left\lVert#1\right\rVert}
\renewcommand{\varepsilon}{\epsilon}
\renewcommand{\tilde}{\wt}
\renewcommand{\hat}{\wh}
\renewcommand{\R}{\mathbb{R}}
\renewcommand{\N}{\mathcal{N}}
\makeatletter
\newcommand*{\rom}[1]{\expandafter\@slowromancap\romannumeral #1@}
\makeatother

\newcommand{\x}{{x}}
\newcommand{\xo}{{x_0}}
\newcommand{\W}[1]{\mathbf{W}^{#1}}
\newcommand{\DD}[1]{\mathbf{D}^{#1}}
\newcommand{\Lam}[1]{\bm{\Lambda}^{#1}}
\newcommand{\upbias}[1]{\mathbf{T}^{#1}}
\newcommand{\lwbias}[1]{\mathbf{H}^{#1}}
\newcommand{\upbnd}[1]{\bm{u}^{#1}}
\newcommand{\lwbnd}[1]{\bm{l}^{#1}}
\newcommand{\z}{\bm{z}}
\newcommand{\y}{\bm{y}}
\newcommand{\bias}[1]{{b}^{#1}}
\newcommand{\setA}{\mathcal{A}}
\newcommand{\setIpos}[1]{\mathcal{I}^{+}_{#1}}
\newcommand{\setIneg}[1]{\mathcal{I}^{-}_{#1}}
\newcommand{\setIuns}[1]{\mathcal{I}_{#1}}
\newcommand{\set}[1]{\mathcal{#1}}
\newcommand{\Lipsloc}{L_{q,x_0}^j}

\newcommand{\gradl}[1]{\tilde{\bm{l}}^{#1}}
\newcommand{\gradu}[1]{\tilde{\bm{u}}^{#1}}
\newcommand{\gradC}[1]{\mathbf{C}^{#1}}
\newcommand{\gradL}[1]{\mathbf{L}^{#1}}
\newcommand{\gradU}[1]{\mathbf{U}^{#1}}
\newcommand{\gradLp}[1]{\mathbf{L'}^{#1}}
\newcommand{\gradUp}[1]{\mathbf{U'}^{#1}}
\newcommand{\Y}[1]{\mathbf{Y}^{#1}}
\newcommand{\Yp}[1]{\mathbf{Y'}^{#1}}

\newcommand{\Au}[1]{\mathbf{\Lambda}^{#1}}
\newcommand{\Al}[1]{\mathbf{\Omega}^{#1}}
\newcommand{\Du}[1]{\mathbf{\lambda}^{#1}}
\newcommand{\Dl}[1]{\mathbf{\omega}^{#1}}
\newcommand{\Ball}{\mathbb{B}_{p}(\xo,\epsilon)}
\newcommand{\Ph}[1]{\Phi_{#1}}
\newcommand{\upslp}[2]{\mathbf{\alpha}^{#1}_{U,{#2}}}
\newcommand{\lwslp}[2]{\mathbf{\alpha}^{#1}_{L,{#2}}}
\newcommand{\upicp}[2]{\mathbf{\beta}^{#1}_{U,{#2}}}
\newcommand{\lwicp}[2]{\mathbf{\beta}^{#1}_{L,{#2}}}

\newcommand{\gradupbnd}[1]{\bm{u'}^{#1}}
\newcommand{\gradlwbnd}[1]{\bm{l'}^{#1}}
\newcommand{\M}[1][]{\mathbf{M}^{#1}}
\newcommand{\setYp}[1]{\mathcal{T}^{+}_{#1}}
\newcommand{\setYn}[1]{\mathcal{T}^{-}_{#1}}
\newcommand{\setYo}[1]{\mathcal{T}_{#1}}
\newcommand{\A}{\mathbf{A}}
\newcommand{\B}{\mathbf{B}}

\newcommand{\hatW}[1]{\widehat{\mathbf{W}}^{#1}}
\newcommand{\checkW}[1]{\widecheck{\mathbf{W}}^{#1}}

\subsection{Overview}
In this section, we present RecurJac, our recursive algorithm for uniformly bounding (local) Jacobian matrix of neural networks with a wide range of activation functions.

\paragraph{Notations.} For an $H$-layer neural network $f(\x)$ with input $x \in \R^{n_0}$, weight matrices $\W{(l)} \in \R^{n_l \times n_{l-1}}$ and bias vectors $\bias{(l)} \in \R^{n_l}$, the network $f(\x)$ can be defined recursively as $h^{(l)}(x)=\sigma^{(l)}(\W{(l)} h^{(l-1)}(x)+b^{(l)})$ for all $l \in \{1,\dots, H-1\}$ with $h^{(0)} \coloneqq x$, $f(x)=\W{(H)} h^{(H-1)} (\x) + \bias{(H)}$. $\sigma^{(l)}$ is a component-wise activation function of (leaky-)ReLU, sigmoid family (including sigmoid, arctan, hyperbolic tangent, etc), and other activation functions that satisfy the assumptions we will formally show below. We denote $\W{(l)}_{r,:}$ as the $r$-th row and $\W{(l)}_{:,j}$ as the $j$-th column of $\W{(l)}$. For convenience, we denote $f^{(l)}(\x) \coloneqq \W{(l)} h^{(l-1)} (\x) + \bias{(l)}$ as the pre-activation function values.

\paragraph{Local Lipchitz constant.} 
Given a function  $f(\x): \R^{n} \rightarrow \R^{m}$ and two distance metrics $d$ and $d'$ on $\R^{n}$ and $\R^{m}$, respectively, the local Lipschitz constant $L^S_{d, d'}$ of $f$ in a close ball of radius $R$ centered at $s$ (denoted as $S = B_{d}[s; R]$) is defined as:
\[
d'(f(x),f(y)) \le L^S_{d, d'} d(x, y), \text{for all $x,y \in S := B_{d}[s; R]$}
\]

Any scalar $L^S_{d, d'}$ that satisfies this condition is a local Lipschitz constant. A good local Lipschitz constant should be as small as possible, {\em i.e.}, close to \textit{the best} (smallest) local Lipschitz constant. A Lipschitz constant we compute can be seen as an upper bound of the best Lipschitz constant.

\paragraph{Assumptions on activation functions.} 
RecurJac has the following assumptions on the activation function $\sigma(x)$:
\begin{assumption} \label{assump:act1}
$\sigma(x)$ is continuous and differentiable almost everywhere on $\R$. This is a basic assumption for neural network activation functions.
\end{assumption}
% \begin{remark} By the extreme value theorem, for a closed interval $[a,b]$, $\sigma$ must attain a maximum and a minimum, each at least once.
% \end{remark}
\begin{assumption} 
\label{assump:act2}
There exists a positive constant $C$ such that $0 \le \sigma'(x) \le C$ when the derivative exists. This covers all common activation functions, including (leaky-)ReLU, hard-sigmoid, exponential linear units (ELU), sigmoid, tanh, arctan and all sigmoid-shaped family activation functions. This assumption helps us derive an elegant bound.
\end{assumption}

\paragraph{Overview of Techniques.} The local Lipschitz constant can be presented as the maximum directional derivative inside the ball $B_{d}[s; R]$~\citep{weng2018evaluating}. For differentiable functions, this is the maximum norm of gradient with respect to the distance metric $d$ (or the maximal operator norm of Jacobian induced by distances $d'$ and $d$ in the vector-output case). 
%Deriving the exact maximum norm of gradient is NP-complete even for a 2-layer ReLU network~\citep{raghunathan2018certified}, so 
We bound each element of Jacobian through a layer-by-layer approach, as shown below.

Define diagonal matrices $\Sigma$ representing the derivatives of the activation functions:
\[
\Sigma^{(l)} := \text{diag}\{\sigma'(f^{(l)}(\x))\}.
\]
The Jacobian matrix of a $H$-layer network can be written as:
\begin{equation}
\label{eq:gradient}
\nabla f^{(H)}(\x) = \W{(H)} \Sigma^{(H-1)} \W{(H-1)} \cdots \W{(2)} \Sigma^{(1)} \W{(1)}.
\end{equation}
For the ease of notation, we also define
\[
\Y{(-l)} := \frac{\partial f^{(H)}}{\partial h^{(l-1)}}=\W{(H)} \Sigma^{(H-1)} \cdots \W{(l+1)} \Sigma^{(l)} \W{(l)}
\]
for $l \in [H]$. As a special case, $\Y{(-1)} := \nabla f^{(H)}$.

In the first step, we assume that we have the following pre-activation bounds $\lwbnd{(l)}_r$ and $\upbnd{(l)}_r$ for every layer $l \in [H-1]$:
\begin{equation}
\label{eq:crown_ul}
\lwbnd{(l)}_r \leq f^{(l)}_r(\x) \leq \upbnd{(l)}_r \quad \forall r\in[n_l], x \in B_{d}[s; R]
\end{equation}
We can get these bounds efficiently via any algorithms that compute layer-wise activation bounds, including CROWN~\citep{zhang2018crown} and convex adversarial polytope~\citep{wong2018provable}. 
%We say a diagonal element in $\Sigma^{(l)}_{r,r}$ is \textit{certain} when we know that $\Sigma^{(l)}_{r,r}$ is fixed to a certain value during our bounding procedure. 
Because pre-activations are within some ranges rather than fixed values, $\Sigma$ matrices contain uncertainties, which will be characterized analytically.

In the second step, we compute both lower and upper bounds for each entry of $\Y{(-l)}:=\frac{\partial f^{(H)}}{\partial h^{(l-1)}}$ in a backward manner. More specifically, we compute $\gradL{(-l)}, \gradU{(-l)} \in \R^{n_H \times n_{l-1}}$ so that
\begin{equation}
\label{eq:gradient_ul}
\gradL{(-l)} \leq \Y{(-l)}(x) \leq \gradU{(-l)} \quad \forall x \in B_{d}[s; R]
\end{equation}
holds true element-wisely. 
For layer $H$, we have $\Y{(-H)}= \W{(H)}$ and thus $\gradL{(-H)} = \gradU{(-H)} = \W{(H)}$. For layers $l< H$, uncertainties in $\Sigma$ matrices propagate into $\Y{(-l)}$ layer by layer. Naively deriving $(\gradL{(-l+1)}, \gradU{(-l+1)})$ just from $(\gradL{(-l)}, \gradU{(-l)})$ and $(\lwbnd{(l-1)}, \upbnd{(l-1)})$ leads to a very loose bound. We propose a fast recursive algorithm that makes use of bounds for all previous layers to compute a much tighter bound for $\Y{(-l)}$. Applying our algorithm to $\Y{(-H+1)}, \Y{(-H+2)}, \cdots$ will eventually allow us to obtain $\Y{(-1)}$. Our algorithm can also be applied in a forward manner; the forward version (RecurJac-F) typically leads to slightly tighter bounds but can slow down the computation significantly, as we will show in the experiments.

From an optimization perspective, we essentially try to solve two constrained maximization and minimization problems with variables $\Sigma^{(l)}_{r,r}$, for each element $\{j,k\}$ in the Jacobian $\nabla f^{(H)}(\x)$:

\begin{align}
\label{eq:opt_lipschitz}
&\max_{\lwbnd{(l)}_r \leq \Sigma^{(l)}_{r,r} \leq \upbnd{(l)}_r} [\nabla f^{(H)}(\x)]_{j,k} \quad \text{and} \quad \min_{\lwbnd{(l)}_r \leq \Sigma^{(l)}_{r,r} \leq \upbnd{(l)}_r} [\nabla f^{(H)}(\x)]_{j,k}
\end{align}

\citet{raghunathan2018certified} show that even for ReLU networks with one hidden layer, finding the maximum $\ell_1$ norm of the gradient is equivalent to the \textsf{Max-Cut} problem and NP-complete. RecurJac is a polynomial time algorithm to give upper and lower bounds on $[\nabla f^{(H)}(\x)]_{j,k}$, rather than solving the exact maxima and minima in exponential time.

After obtaining the Jacobian bounds $\Y{(-1)}:= \nabla f^{(H)}$, we can make use of it to derive an upper bound for the local Lipchitiz constant in the set $S = B_{d}[s; R]$. We present bounds when $d$ and $d'$ are both ordinary $p$-norm ($p=[1,+\infty) \cup \{+\infty\}$) distance in Euclidean space. We can also use the Jacobian bounds for other proposes, like understanding the local optimization landscape.

\subsection{Bounds for $\Sigma^{(l)}$}
\label{subsec:sigma_bnd}
From \eqref{eq:gradient}, we can see that the uncertainties in $\nabla f^{(H)}$ are purely from $\sigma'(f^{(l)}(\x))$; all $\W{(l)}$ are fixed. For any $l \in [H-1]$, we define the range of $\sigma'(f^{(l)}_r(x))$ as $\gradlwbnd{(l)}_r$ and $\gradupbnd{(l)}_r$, i.e.,
\begin{equation}
\label{eq:sigma_bnd}
\gradlwbnd{(l)}_r \leq \sigma'(f^{(l)}_r(x)) \leq \gradupbnd{(l)}_r \quad \forall r \in [n_{l}].
\end{equation}
Note that $\gradlwbnd{(l)}_r$ and $\gradupbnd{(l)}_r$ can be easily obtained because we know $\lwbnd{(l)}_r \le f^{(l)}_r(x) \le \upbnd{(l)}_r$ (thanks to \eqref{eq:crown_ul}) and the analytical form of $\sigma'(x)$. For example, for the sigmoid function $\sigma(x) = \frac{e^x}{1 + e^x}$, $\sigma'(x) = \sigma(x)(1-\sigma(x))$, we have:

\begin{align}
\label{eq:grad_lr}
\gradlwbnd{(l)}_r & = 
    \begin{cases}
    \sigma'(\lwbnd{(l)}_r) & \text{if $\lwbnd{(l)}_r \leq \upbnd{(l)}_r \leq 0$;} \\
    \sigma'(\upbnd{(l)}_r) & \text{if $\upbnd{(l)}_r \geq \lwbnd{(l)}_r \geq 0$;} \\
    \sigma'( \max\{-\lwbnd{(l)}_r, \upbnd{(l)}_r\}) & \text{if $\lwbnd{(l)}_r \leq 0 \leq \upbnd{(l)}_r$.}
  \end{cases}
\end{align}

\begin{align}
\label{eq:grad_ur}
\gradupbnd{(l)}_r & = 
    \begin{cases}
    \sigma'(\upbnd{(l)}_r) & \text{if $\lwbnd{(l)}_r \leq \upbnd{(l)}_r \leq 0$;} \\
    \sigma'(\lwbnd{(l)}_r) & \text{if $\upbnd{(l)}_r \geq \lwbnd{(l)}_r \geq 0$;} \\
    \sigma'(0) & \text{if $\lwbnd{(l)}_r \leq 0 \leq \upbnd{(l)}_r$.}
  \end{cases}
\end{align}

Equations~\eqref{eq:grad_lr} and \eqref{eq:grad_ur} are also valid for other sigmoid-family activation functions, including $\sigma(x)=\frac{e^x}{1 + e^x}$, $\sigma(x)=\tanh(x)$, $\sigma(x)=\arctan(x)$ and many others.

For (leaky-)ReLU activation functions with a negative-side slope $\alpha$ ($0 \le \alpha \le 1$), $\gradlwbnd{(l)}_r$ and $\gradupbnd{(l)}_r$ are:

\begin{align*}
% \label{eq:grad_lr_relu}
\gradlwbnd{(l)}_r & = 
    \begin{cases}
    \alpha & \text{if $\lwbnd{(l)}_r \leq \upbnd{(l)}_r \leq 0$ \enskip or \enskip $\lwbnd{(l)}_r \leq 0 \leq \upbnd{(l)}_r$;} \\
    1 & \text{if $\upbnd{(l)}_r \geq \lwbnd{(l)}_r \geq 0$.}
  \end{cases}
\end{align*}

\begin{align*}
% \label{eq:grad_ur_relu}
\gradupbnd{(l)}_r & = 
    \begin{cases}
    \alpha & \text{if $\lwbnd{(l)}_r
    \leq \upbnd{(l)}_r \leq 0$;} \\
    1 & \text{if $\upbnd{(l)}_r \geq \lwbnd{(l)}_r \geq 0$ \enskip or \enskip $\lwbnd{(l)}_r \leq 0 \leq \upbnd{(l)}_r$.}
  \end{cases}
\end{align*}

For (leaky-)ReLU activation functions, in the cases where $\lwbnd{(l)}_r \leq \upbnd{(l)}_r \leq 0$ and $\upbnd{(l)}_r \geq \lwbnd{(l)}_r \geq 0$, we have $\gradlwbnd{(l)}_r = \gradupbnd{(l)}_r$, so $\Sigma^{(l)}_{r,r}$ becomes a constant and there is no uncertainty.

\subsection{A recursive algorithm to bound $\Y{(-l)}$} 
\label{subsec:recursive_alg}

\paragraph{Bounds for $\Y{(-H+1)}$.}
By definition, we have $\Y{(-H)} = \W{(H)}$ and $\Y{(-H+1)} = \Y{(-H)} \Sigma^{(H-1)} \W{(H-1)}$. Thus,
\begin{equation}
\label{eq:lips_2layer}
\Y{(-H+1)}_{j,k} = \sum_{r \in [n_{H-1}]} \W{(H)}_{j,r} \sigma'(f^{(H-1)}_r) \W{(H-1)}_{r,k},
\end{equation}
where $\gradlwbnd{(H-1)}_r \le \sigma'(f^{(H-1)}_r) \le \gradupbnd{(H-1)}_r$ thanks to \eqref{eq:sigma_bnd}.

By assumption~\ref{assump:act2}, $\sigma'(x)$ is always non-negative, and thus we only need to consider the signs of $\W{(H)}$ and $\W{(H-1)}$. Denote $\gradL{(-H+1)}_{j,k}$ and $\gradU{(-H+1)}_{j,k}$ to be a lower and upper bounds of~\eqref{eq:lips_2layer}. By examining the signs of each term, we have

\begin{align}
\begin{split}
\label{eq:2layer_lb}
\gradL{(-H+1)}_{j,k} &= \hspace{-4mm} \sum_{\W{(H)}_{j,r}\W{(H-1)}_{r,k} < 0} \hspace{-6mm} \gradupbnd{(H-1)}_r \W{(H)}_{j,r}\W{(H-1)}_{r,k} 
%\\ & 
+ \sum_{\W{(H)}_{j,r}\W{(H-1)}_{r,k} > 0} \hspace{-6mm} \gradlwbnd{(H-1)}_r \W{(H)}_{j,r}\W{(H-1)}_{r,k} , 
\end{split}
\end{align}

\begin{align}
\begin{split}
\label{eq:2layer_ub}
\gradU{(-H+1)}_{j,k} &= \hspace{-4mm} \sum_{\W{(H)}_{j,r}\W{(H-1)}_{r,k} > 0} \hspace{-6mm} \gradupbnd{(H-1)}_r \W{(H)}_{j,r}\W{(H-1)}_{r,k} 
%\\ & 
+ \sum_{\W{(H)}_{j,r}\W{(H-1)}_{r,k} < 0} \hspace{-6mm} \gradlwbnd{(H-1)}_r \W{(H)}_{j,r}\W{(H-1)}_{r,k} .
\end{split}
\end{align}

In \eqref{eq:2layer_lb}, we collect all negative terms of $\W{(H)}_{j,r}\W{(H-1)}_{r,k}$ and multiply them by $\gradupbnd{(H-1)}_r$ as a lower bound of $\sum\limits_{\W{(H)}_{j,r}\W{(H-1)}_{r,k} < 0} \sigma'(f_r^{(H-1)}(x)) \W{(H)}_{j,r}\W{(H-1)}_{r,k}$, and collect all positive terms and multiply them by $\gradlwbnd{(H-1)}_r$ as a lower bound of the positive counterpart. We obtain the upper bound in \eqref{eq:2layer_ub} following the same rationale. Fast-Lip is a special case of RecurJac when there are only two layers with ReLU activations; RecurJac becomes much more sophisticated in multi-layer cases, as we will show below.

\paragraph{Bounds for $\Y{(-l)}$ when $1 \leq l < H-1$.}
By definition, we have $\Y{(-l+1)} = \Y{(-l)} \Sigma^{(l-1)} \W{(l-1)}$, i.e.,
\begin{equation}
\label{eq:multilayer}
\Y{(-l+1)}_{j,k} = \sum_{r \in [n_{l-1}]} \Y{(-l)}_{j,r} \sigma'(f_r^{(l-1)}(\x)) \W{(l-1)}_{r,k},
\end{equation}
where $\gradlwbnd{(l-1)}_r \leq \sigma'(f^{(l-1)}_r(x)) \leq \gradupbnd{(l-1)}_r$ thanks to \eqref{eq:sigma_bnd} and 
\[
\gradL{(-l)}_{j,r} \leq \Y{(-l)}_{j,r} \leq \gradU{(-l)}_{j,r} \quad \forall j,r
\]
thanks to previous computation. We want to find the bounds
\[
\gradL{(-l+1)}_{j,k} \leq \Y{(-l+1)}_{j,k} \leq \gradU{(-l+1)}_{j,k} \quad \forall j,k.
\]

We decompose \eqref{eq:multilayer} into two terms:
\begin{align}
\begin{split}
\label{eq:multilayer_decom}
& \Y{(-l+1)}_{j,k} = \underbrace{ \sum_{\{r ~:~ \gradL{(-l)}_{j,r} <0 < \gradU{(-l)}_{j,r}\}} \Y{(-l)}_{j,r}\sigma'(f_r^{(l-1)}(\x)) \W{(l-1)}_{r,k} }_{I} \\ 
& + \underbrace{ \sum_{\{r ~:~ \gradL{(-l)}_{j,r} \ge 0 \text{ or } \gradU{(-l)}_{j,r} \le 0\}} \Y{(-l)}_{j,r}\sigma'(f_r^{(l-1)}(\x)) \W{(l-1)}_{r,k} }_{II},
\end{split}
\end{align}
and bound them separately.

Observing the signs of each term in $I$ and $\gradupbnd{(l+1)}_r \geq \gradlwbnd{(l+1)}_r \geq 0$, we take:

\begin{align}
\gradL{(-l+1),\pm}_{j,k} \hspace{-1mm} &= \hspace{-4mm} \sum_{\W{(l-1)}_{r,k} < 0} \hspace{-4mm} \gradupbnd{(l-1)}_r \gradU{(-l)}_{j,r} \W{(l-1)}_{r,k} %\nonumber \\ & 
\label{eq:nlayer_lb_I}
+ \hspace{-4mm} \sum_{\W{(l-1)}_{r,k} > 0} \hspace{-4mm} \gradupbnd{(l-1)}_r \gradL{(-l)}_{j,r} \W{(l-1)}_{r,k} 
\end{align}
\begin{align}
\gradU{(-l+1),\pm}_{j,k} \hspace{-1mm} &= \hspace{-4mm} \sum_{\W{(l-1)}_{r,k} < 0} \hspace{-4mm} \gradupbnd{(l-1)}_r \gradL{(-l)}_{j,r} \W{(l-1)}_{r,k} %\nonumber \\ &
\label{eq:nlayer_ub_I}
+ \hspace{-4mm} \sum_{\W{(l-1)}_{r,k} > 0} \hspace{-4mm} \gradupbnd{(l-1)}_r \gradU{(-l)}_{j,r} \W{(l-1)}_{r,k}
\end{align}
The index constraint $\{r ~:~ \gradL{(-l)}_{j,r} <0< \gradU{(-l)}_{j,r}\}$ is still effective in \eqref{eq:nlayer_lb_I} and \eqref{eq:nlayer_ub_I}, but we omit it for notation simplicity. Then we can show that $\gradL{(-l+1),\pm}_{j,k}$ and $\gradU{(l+1),\pm}_{j,k}$ are a lower and upper bound for term $I$ in \eqref{eq:multilayer_decom} as follows.

\begin{proposition}
\label{prop:recursion_I}
\begin{equation}
\label{eq:recursion_I}
\gradL{(-l+1),\pm}_{j,k} \leq I \leq \gradU{(-l+1),\pm}_{j,k},
\end{equation}
where $I$ is the first term in \eqref{eq:multilayer_decom}.
\end{proposition}

For term $II$ in \eqref{eq:multilayer_decom}, the sign of $\Y{(-l)}_{j,r}$ does not change since $\gradL{(-l)}_{j,r} \ge 0$ or $\gradU{(-l)}_{j,r} \le 0$.  Similar to what we did in \eqref{eq:2layer_lb} and \eqref{eq:2layer_ub}, depending on the sign of $\Y{(-l)}_{j,r} \W{(l-1)}_{r,k}$, we can lower/upper bound term $II$ using $\Y{(-l)}$ itself instead of its bound $(\gradL{(-l)}, \gradU{(-l)})$. This will give us much tighter bounds than just naively using $(\gradL{(-l)}, \gradU{(-l)})$ as we deal with term $I$. More specifically, we define $2 n_H$ matrices $\checkW{(l,l-1,j)}, \hatW{(l,l-1,j)} \in \R^{n_{l}\times n_{l-2}}$ for $j \in [n_H]$ as below: 
\begin{align}
\begin{split}
\label{eq:multilayer_lb_II}
\checkW{(l,l-1,j)}_{i,k} &= \hspace{-4mm} \sum_{\substack{\gradL{(-l)}_{j,r} \ge 0, \W{(l-1)}_{r,k}>0 \\ \text{or } \gradU{(-l)}_{j,r} \le 0, \W{(l-1)}_{r,k}<0}} \W{(l)}_{i,r} \gradlwbnd{(l-1)}_r \W{(l-1)}_{r,k} 
% \\ & 
+ \hspace{-4mm} \sum_{\substack{\gradL{(-l)}_{j,r} \ge 0, \W{(l-1)}_{r,k}<0 \\ \text{or } \gradU{(-l)}_{j,r} \le 0, \W{(l-1)}_{r,k}>0}} \W{(l)}_{i,r} \gradupbnd{(l-1)}_r \W{(l-1)}_{r,k},
\end{split}
\end{align}

\begin{align}
\begin{split}
\label{eq:multilayer_ub_II}
\hatW{(l,l-1,j)}_{i,k} &= \hspace{-4mm} \sum_{\substack{\gradL{(-l)}_{j,r} \ge 0, \W{(l-1)}_{r,k}>0 \\ \text{or } \gradU{(-l)}_{j,r} \le 0, \W{(l-1)}_{r,k}<0}} \W{(l)}_{i,r} \gradupbnd{(l-1)}_r \W{(l-1)}_{r,k} 
% \\ & 
+ \hspace{-4mm} \sum_{\substack{\gradL{(-l)}_{j,r} \ge 0, \W{(l-1)}_{r,k}<0 \\ \text{or } \gradU{(-l)}_{j,r} \le 0, \W{(l-1)}_{r,k}>0}} \W{(l)}_{i,r} \gradlwbnd{(l-1)}_r \W{(l-1)}_{r,k}.
\end{split}
\end{align}
Then we can show the following lemma.

\begin{lemma}
\label{lem:recursion_II}
For any $j \in [n_H]$, we have
\begin{equation}
\label{eq:recursion_II}
\Y{(-l-1)}_{j,:} \Sigma^{(l)} \checkW{(l,l-1,j)}_{:,k} \leq II \leq \Y{(-l-1)}_{j,:} \Sigma^{(l)} \hatW{(l,l-1,j)}_{:,k},
\end{equation}
where $II$ is the second term in \eqref{eq:multilayer_decom}.
\end{lemma}
Note that when the sign of $\Y{(-l)}_{j,r}$ is fixed, i.e., $\gradL{(-l)}_{j,r} \ge 0$ or $\gradU{(-l)}_{j,r} \le 0$ in term $II$, the bounds in \eqref{eq:recursion_II} is always tighter than those in \eqref{eq:recursion_I}. After we know the sign of $\Y{(-l)}_{j,r}$, we can fix $\sigma'(f_r^{(l-1)}(\x))$ to be either $\gradlwbnd{(l-1)}_r$ or $\gradupbnd{(l-1)}_r$ according to the sign of $\W{(l-1)}_{r,k}$ and thus eliminate the uncertainty in $\sigma'(f_r^{(l-1)}(\x))$. Then we can plug $\Y{(-l)}_{j,r} = \sum_{s} \Y{(-l-1)}_{j,i} \sigma'(f_i^{(l)}(\x)) \W{(l)}_{i,r}$ into the lower and upper bounds and merge terms involving $\W{(l)}_{i,r}$, $\sigma'(f_r^{(l-1)}(\x))$ and $\W{(l-1)}_{r,k}$, resulting in~\eqref{eq:recursion_II}. Compared with using the worst-case bound $\gradL{(-l)}_{j,r} \leq \Y{(-l)}_{j,r} \leq \gradU{(-l)}_{j,r}$ directly in \eqref{eq:recursion_I}, we expand $\Y{(-l)}_{j,r}$ and remove uncertainty in $\sigma'(f_r^{(l-1)}(\x))$ in \eqref{eq:recursion_II}, and thus get much tighter bounds.

% Note that in~\eqref{eq:multilayer} we have $\Y{(-l)}_{j,r}$ multiplies $\W{(l-1)}_{r,k}$, but now we have $\Y{(-l-1)}_{j:,}$ multiplies $\checkW{}$ or $\hatW{}$. $\Y{(-l)}$ has been eliminated in II because after we know the sign of $\Y{(-l)}_{j,r}$, we can fix $\sigma'(f_r^{(l-1)}(\x))$ to be either $\gradlwbnd{(l-1)}_r$ or $\gradupbnd{(l-1)}_r$ according to the sign of $\W{(l-1)}_{r,k}$, as shown in~\eqref{eq:multilayer_lb_II} and \eqref{eq:multilayer_ub_II}. This reduces uncertainty in $\Sigma^{(l-1)}$ and thus improve the bound.

Finally, combining Proposition~\ref{prop:recursion_I} and Lemma~\ref{lem:recursion_II}, we get the following recursive formula to bound $\Y{(-l+1)}$.

\begin{theorem}
\label{thm:multiple_LU}
For any $1 < l < H$ and any $j\in [n_H]$, we have
\[
\Y{(-l+1)}_{j,:} \ge \gradL{(-l+1),\pm}_{j,:} + \Y{(-l-1)}_{j,:} \Sigma^{(l)} \checkW{(l,l-1,j)}_{:,k}
\]
and 
\[
\Y{(-l+1)}_{j,:} \le \gradU{(-l+1),\pm}_{j,:} + \Y{(-l-1)}_{j,:} \Sigma^{(l)} \hatW{(l,l-1,j)}_{:,k},
\]
where $\gradL{(-l+1),\pm}, \gradU{(-l+1),\pm}, \checkW{(l,l-1,j)}$ and $\hatW{(l,l-1,j)}$ are defined in \eqref{eq:nlayer_lb_I}, \eqref{eq:nlayer_ub_I}, \eqref{eq:multilayer_lb_II} and \eqref{eq:multilayer_ub_II}, respectively. 
\end{theorem}

\begin{remark} The lower and upper bounds of $\Y{(-H+1)}$ in \eqref{eq:2layer_lb} and \eqref{eq:2layer_ub} can be viewed as a special case of Theorem~\ref{thm:multiple_LU} when $l = H$. Because we have $\gradL{(-H)}=\gradU{(-H)}=\W{(l)}$ in this case, we do not have term $I$ in the decomposition \eqref{eq:multilayer_decom}. Moreover, the bounds of term $II$ in \eqref{eq:recursion_II} are reduced to exactly \eqref{eq:2layer_lb} and \eqref{eq:2layer_ub} after we notice that $\checkW{(H,H-1,j)}_{j,k} = \gradL{(-H+1)}_{j,k}$ and $\hatW{(H,H-1,j)}_{j,k} = \gradU{(-H+1)}_{j,k}$ and specify $\Y{(-H-1)}=\Sigma^{(H)}=I_{n_H}$. Specifying $\Y{(-H-1)}=\Sigma^{(H)}=I_{n_H}$ is equivalent to adding another identity layer to the neural network $f^{(H)}(x)$, which does not change any computation.
\end{remark}

\paragraph{A recursive algorithm to bound $\Y{(-l)}$.} 
Notice that the lower and upper bounds in Lemma~\ref{lem:recursion_II} have exactly the same formation of $\Y{(-l)} = \Y{(-l-1)} \Sigma^{(l)} \W{(l)}$, by replacing $\W{(l)}$ with $\checkW{(l,l-1,j)}$ and $\hatW{(l,l-1,j)}$. Therefore, we can recursively apply our Theorem~\ref{thm:multiple_LU} to obtain an lower and upper bound for $\Y{(-l+1)}$, denoted as $\gradL{(-l+1)}$ and $\gradU{(-l+1)}$ separately. This recursive procedure further reduces uncertainty in $\Sigma$ for all previous layers, improving the quality of bounds significantly. We elaborate our recursive algorithm in Algorithm~\ref{alg:recursive_LU} for the case $n_H=1$, so we omit the last superscript $j=1$ in $\checkW{(l,l-1,1)}$ and $\hatW{(l,l-1,1)}$. When $n_H>1$, we can apply Algorithm~\ref{alg:recursive_LU} independently for each output.

\begin{algorithm}
% \caption{Compute $\gradL{(-l)}$ and $\gradU{(-l)}$ for $1 \le l \le H$ when $n_H=1$} 
% \caption{ ($\gradL{(-l)}$, $\gradU{(-l)}$) = ComputeLU( $\W{(l)}$, $\{(\gradL{(-i)}, \gradU{(-i)}, \W{(i)})\}_{i=l+1}^H$,~ $\{\gradlwbnd{(i-1)}, \gradupbnd{(i-1)}\}_{i=l+1}^H$) }
\caption{ComputeLU (compute the lower and upper Jacobian bounds)}
% \begin{algorithmic}
% $(\gradL{(-l)}$, $\gradU{(-l)})$ = ComputeLU($\{(\gradL{(-i)}, \gradU{(-i)}, \W{(i)})\}_{i=l+1}^H$, $\{\gradlwbnd{(i-1)}, \gradupbnd{(i-1)}\}_{i=l+1}^H$, $\W{(l)}$)
\begin{algorithmic}[1]
    \Require{$\W{(l)}$, bounds $\{(\gradL{(-i)}, \gradU{(-i)}, \W{(i)})\}_{i=l+1}^H$, $\{\gradlwbnd{(i-1)}, \gradupbnd{(i-1)}\}_{i=l+1}^H$}
    \If{$l=H$}
        \State $\gradL{(-l)}=\gradU{(-l)}=\W{(l)}$
    \ElsIf{$l=H-1$}
        \State Compute $\gradL{(-l)}$ from \eqref{eq:2layer_lb}, $\gradU{(-l)}$ from \eqref{eq:2layer_ub}
    \ElsIf{$1 \le l < H-1$}
        \State Compute $\checkW{(l+1,l)}$ from \eqref{eq:multilayer_lb_II}, $\hatW{(l+1,l)}$ from \eqref{eq:multilayer_ub_II}
        \State $(\gradL{(-l-1,-l)}\!,\! \backsim)$ \!=\! ComputeLU( $\checkW{(l+1,l)}$, $\{(\gradL{(-i)}\!,\! \gradU{(-i)}\!,\! \W{(i)})\}_{i=l+2}^H$,~ $\{\gradlwbnd{(i-1)}, \gradupbnd{(i-1)}\}_{i=l+2}^H$) \\ \Comment{Recursive call}
        \State $(\backsim\!,\! \gradU{(-l-1,-l)})$ \!=\! ComputeLU( $\hatW{(l+1,l)}$, $\{(\gradL{(-i)}\!,\! \gradU{(-i)}\!,\! \W{(i)})\}_{i=l+2}^H$,~ $\{\gradlwbnd{(i-1)}, \gradupbnd{(i-1)}\}_{i=l+2}^H$)\\
        \Comment{Recursive call}
        \State Compute $\gradL{(-l),\pm}$ from \eqref{eq:nlayer_lb_I}, $\gradU{(-l),\pm}$ from \eqref{eq:nlayer_ub_I}
        \State $\gradL{(-l)} = \gradL{(-l),\pm} + \gradL{(-l-1,-l)}$
        \State $\gradU{(-l)} = \gradU{(-l),\pm} + \gradU{(-l-1,-l)}$
    \EndIf
    \State \Return $\gradL{(-l)}$, $\gradU{(-l)}$
\end{algorithmic}
\label{alg:recursive_LU}
\end{algorithm}

\paragraph{Compute the bounds in a forward manner.} 
In previous sections, we start our computation from the last layer and bound $\Y{(-l)}:=\frac{\partial f^{(H)}}{\partial h^{(l-1)}}$ in a backward manner. By transposing \eqref{eq:gradient}, we have
\[
[ \nabla f^{(H)}(\x) ]^T = \W{(1)T} \Sigma^{(1)} \W{(2)T} \cdots \Sigma^{(H-1)} \W{(H)T}.
\]
Then we can apply Algorithm~\ref{alg:recursive_LU} to bound $\nabla f^{(H)}(\x)^T$ according to the equation above. This is equivalent to starting from the first layer, and bound $\frac{\partial f^{(l)}}{\partial x}$ from $l=1$ to $H$. Because we obtain the bounds of pre-activations in a forward manner by CROWN~\citep{zhang2018crown}, the bounds \eqref{eq:crown_ul} get looser when the layer index $l$ gets larger. Therefore, bounding the Jacobian by the forward version is expected to get tighter bounds of $\frac{\partial f^{(l)}}{\partial x}$ at least for small $l$. In our experiments, we see that the bounds for $\nabla f^{(H)}(\x)$ obtained from the forward version are typically a little tighter than those obtained from the backward version. However, the ``output'' dimension in this case is $n_0$, which is the input dimension of the neural network. For image classification networks, $n_H \ll n_0$, the forward version has to apply Algorithm~\ref{alg:recursive_LU} $n_0$ times to obtain the final bounds and thus increases the computational cost significantly compared to the backward version. We make a detailed comparison between the forward and backward version in the experiment section. 

\subsection{Compute a local Lipschitz constant}
\label{sec:localLip_alg}
After obtaining $\gradL{(-1)} \le \Y{(-1)} := \nabla f(\x) \le \gradU{(-1)}$ for all $\x \in S$, we define
\begin{equation}\label{eq:def_M}
\max_{\x \in S} |[\nabla f(\x)]| \leq  \M \coloneqq \max (|\gradL{(-1)}|, |\gradU{(-1)}|),
\end{equation}
where the $\max$ and inequality are taken element-wise. In the rest of this subsection, we simplify the notations $\Y{(-1)}, \gradL{(-1)}, \gradU{(-1)}$ to $\Y{}, \gradL{}, \gradU{}$ when no confusion arises.

Recall that the Local Lipschitz constant $L^S_d$ can be evaluated as $L^S_{d,d'} = \max_{\x \in S} \| \nabla f(\x) \|_{d,d'}$.
$\nabla f(x)$ is the Jacobian matrix and $\| \cdot \|_{d,d'}$ denotes the induced operator norm.
Then we can bound the maximum norm of Jacobian (local Lipschitz constant) considering its element-wise worst case. When $d$, $d'$ are both ordinary $p$-norm ($p=[1,+\infty) \cup \{+\infty\}$) distance in Euclidean space, we denote $L_{d,d'}$ as $L_p$, and it can be bounded as follows.

\begin{proposition}\label{prop:Lp_bnd}
For any $1 \le p \le +\infty$, we have
\begin{equation}\label{eq:Lp_bnd}
    L^S_{p} := \max_{x \in B_{\ell_p}[s; R]} \| \nabla f(x) \|_{p}  \leq \| \M \|_{p},
\end{equation}
where $\M\coloneqq \max (|\gradL{}|, |\gradU{}|)$ is defined in \eqref{eq:def_M}. 
\end{proposition}

\paragraph{Improve the bound for $L^S_\infty$.}

\begin{algorithm}[b]
\caption{Upper bound of $\max_{x \in B_{\ell_\infty}[s; R]} \| \nabla f(x) \|_{\infty}$} 
\begin{algorithmic}[1]
    \State Compute $\M$ from \eqref{eq:def_M}
    \For{$j \in [n_H]$}
        \State Compute $\hat{w}^{(j)}_r$ from \eqref{eq:def_wj}
        \State $(\backsim, \gradU{(0,j)})$ = ComputeLU( $\hat{w}^{(j)}_r$, $\{(\gradL{(-i)}, \gradU{(-i)}, \W{(i)})\}_{i=1}^H$, $\{\gradlwbnd{(i-1)},~ \gradupbnd{(i-1)}\}_{i=1}^H$)
        \State $\gradU{(0)}_j = \gradU{(0,j)} + \sum_{k \in \setYo{j}} \M_{j,k}$
    \EndFor
    \State \Return $\max_{j\in [n_H]} \gradU{(0)}_j$
\end{algorithmic}
\label{alg:Linf_U}
\end{algorithm}

For the important case of upper bounding $L^S_\infty$, we use an additional trick to improve the bound~\eqref{eq:Lp_bnd}. We note that $\|\nabla f(x)\|_{\infty} = \max_j \sum_{k} | \Y{}_{j,k} |$. As in \eqref{eq:multilayer_decom}, we decompose it into two terms
\begin{equation}\label{eq:Linf_decom}
    \sum_k |\Y{}_{j,k}| = \underbrace{\sum_{k \in \setYo{j}} |\Y{}_{j,k}|}_{I} + \underbrace{\sum_{k \in \setYp{j}} {\Y{}_{j,k}} - \sum_{k \in \setYn{j}} {\Y{}_{j, k}}}_{II}, 
\end{equation}
where $\setYp{j} \coloneqq \{k | \gradL{}_{j,k} \ge 0 \}$, $\setYn{j} \coloneqq \{k | \gradU{}_{j,k} \le 0 \}$, and $\setYo{j} \coloneqq \{k | \gradL{}_{j,k} < 0 < \gradU{}_{j,k}\}$.

For term $I$, we take the same bound as we have in \eqref{eq:Lp_bnd}, i.e., $I \le \sum_{k \in \setYo{j}} \M_{j,k}$.

For term $II$, thanks to $\Y{} = \Y{(-2)}\Sigma^{(1)}\W{(1)}$, we have
\begin{align*}
II &= \sum_{k \in \setYp{j}} {\sum_{r} \Y{(-2)}_{j,r} \sigma'(f_r^{(1)} (x)) \W{(1)}_{r,k}} 
%\\ &
    - \sum_{k \in \setYn{j}} {\sum_{r} \Y{(-2)}_{j,r} \sigma'(f_r^{(1)} (x)) \W{(1)}_{r,k}} \\
    &= \sum_{r} \Y{(-2)}_{j,r} \sigma'(f_r^{(1)} (x)) (\sum_{k \in \setYp{j}} \W{(1)}_{r,k} - \sum_{k \in \setYn{j}} \W{(1)}_{r,k})
\end{align*}

Define $\hat{w}^{(j)} \in \R^{n_1 \times 1}$ and 
\begin{equation}\label{eq:def_wj}
\hat{w}^{(j)}_r \coloneqq \sum_{k \in \setYp{j}} \W{(1)}_{r,k} - \sum_{k \in \setYn{j}} \W{(1)}_{r,k}.
\end{equation}

Combining upper bounds for both terms, we obtain
\[
\sum_j |\Y{}_{i,j}| \le \sum_{k \in \setYo{j}} \M_{j,k} + \Y{(-2)}_{j,:}\Sigma^{(1)}\hat{w}^{(j)}
\]
In the same flavor with Theorem~\ref{thm:multiple_LU}, this bound avoids the worst case bound $\M_{j,k}$ for entries whose signs are known. Notice that $\Y{(-2)}_{j,:}\Sigma^{(1)}\hat{w}^{(j)}$ has exactly the same formation of $\Y{(-1)}$ and we can call Algorithm~\ref{alg:recursive_LU} to get its upper bound.

Finally, assume that from Algorithm~\ref{alg:recursive_LU} we already obtained $\{(\gradL{(-l)}, \gradU{(-l)})\}_{l=1}^H$, we summarize the algorithm to compute upper bound of $L^S_\infty$ in Algorithm~\ref{alg:Linf_U}.

\paragraph{Improve the bound for robustness verification.} In some applications (e.g., robustness verification), we only need to bound $\| f(x) - f(s)\|$ for a fixed $s$ and $x \in B[s; R]$. Although $L^{B[s; R]} \cdot R$ gives a bound of $\| f(x) - f(s)\|$, we can make this bound tighter by using an integral:

\begin{theorem}
\label{thm:robustness_integral}
\[
\| f(x) - f(s) \| \leq \int_0^R L^{B[s; t]} dt \leq L^{B[s; R]} \cdot R, \forall x \in B[s; R].
\]
\end{theorem}

In practice, the integral $\int_0^R L^{B[s; t]} dt$ can be upper bounded by evaluating at $n$ intervals:

\begin{equation}
\label{eq:approx_lip_integral}
\int_0^R L^{B[s; t]} dt \leq \sum_{i=1}^n L^{B[s; t_{i}]} \Delta t 
\end{equation}

where we divide $R$ into $n$ segments $t_0=0, t_1, t_2, \cdots, t_{n-1}, t_n=R$, and $t_{i+1} - t_i = \Delta t$.

\subsection{Extension beyond element-wise activation}

In our theoretical discussion, we define the network as affine transformations plus element-wise activation functions, which cover most elements in modern neural networks (convolutional layers, batch normalization, average pooling, etc), as these operations can be equivalently written $h^{(l)}(x)=\sigma^{(l)}(\W{(l)} h^{(l-1)}(x)+b^{(l)})$ with specially chosen $\W{}$ and $b$. However, there exist network components that perform non-linearity on multiple elements -- notably, max-pooling can be viewed as an activation function applied on multiple neurons, e.g., $\maxp(h_1, \cdots, h_n) = \max(h_1, \cdots, h_n)$, where hidden neurons $h_1, \cdots, h_n$ are within one filter region of max-pooling.

We briefly show that how to incorporate max-pooling in our framework by converting it to a few equivalent layers in the form of $h^{(l)}(x)=\sigma^{(l)}(\W{(l)} h^{(l-1)}(x)+b^{(l)})$. In the simplest case, for a hidden layer $h$ with 2 neurons, we have:

\begin{align*}
\max(h_1, h_2) &= \max(h_1 - h_2, 0) + h_2 = \sigma(h_1 - h_2) + h_2 \\
&= \sigma(h_1 - h_2) + \sigma(h_2) - \sigma(-h_2)
\end{align*}
where $\sigma$ is the element-wise ReLU activation function.
Thus, for $A = \bigl[\begin{smallmatrix}
1&0&0 \\ -1&1&-1
\end{smallmatrix} \bigr]^T, B = [1, 1, -1]^T$, we have $\maxp(h) = B \sigma(A h)$, which can be included in our framework. For a $2\times2$ max-pooling, we have
\begin{align*}
\maxp(h) &=\max(h_1,h_2,h_3,h_4) \\
&=\max(\max(h_1,h_2), \max(h_3,h_4))
\end{align*}

We can then mechanically apply the case for two neurons twice. In general, for max pooling over a group of $n$ (usually $2\times2$, $3\times3$, etc) neurons, we need $O(\log n)$ auxiliary layers to realize max-pooling in our framework. This procedure can also be applied to other works such as CROWN~\citep{zhang2018crown} and convex adversarial polytope~\citep{wong2018provable} to cover max-pooling for their algorithms.

\section{Applications and Experiments}

To demonstrate the effectiveness of RecurJac, we apply it to a variety of networks with different depths, hidden neuron sizes, activation functions, and inputs bounded by different $\ell_p$ norms. Our source code is publicly available\footnote{\url{http://github.com/huanzhang12/RecurJac-Jacobian-bounds}}.

\subsection{Local optimization landscape}

\begin{figure}[ht]
  \centering
  \includegraphics[width=0.33\textwidth]{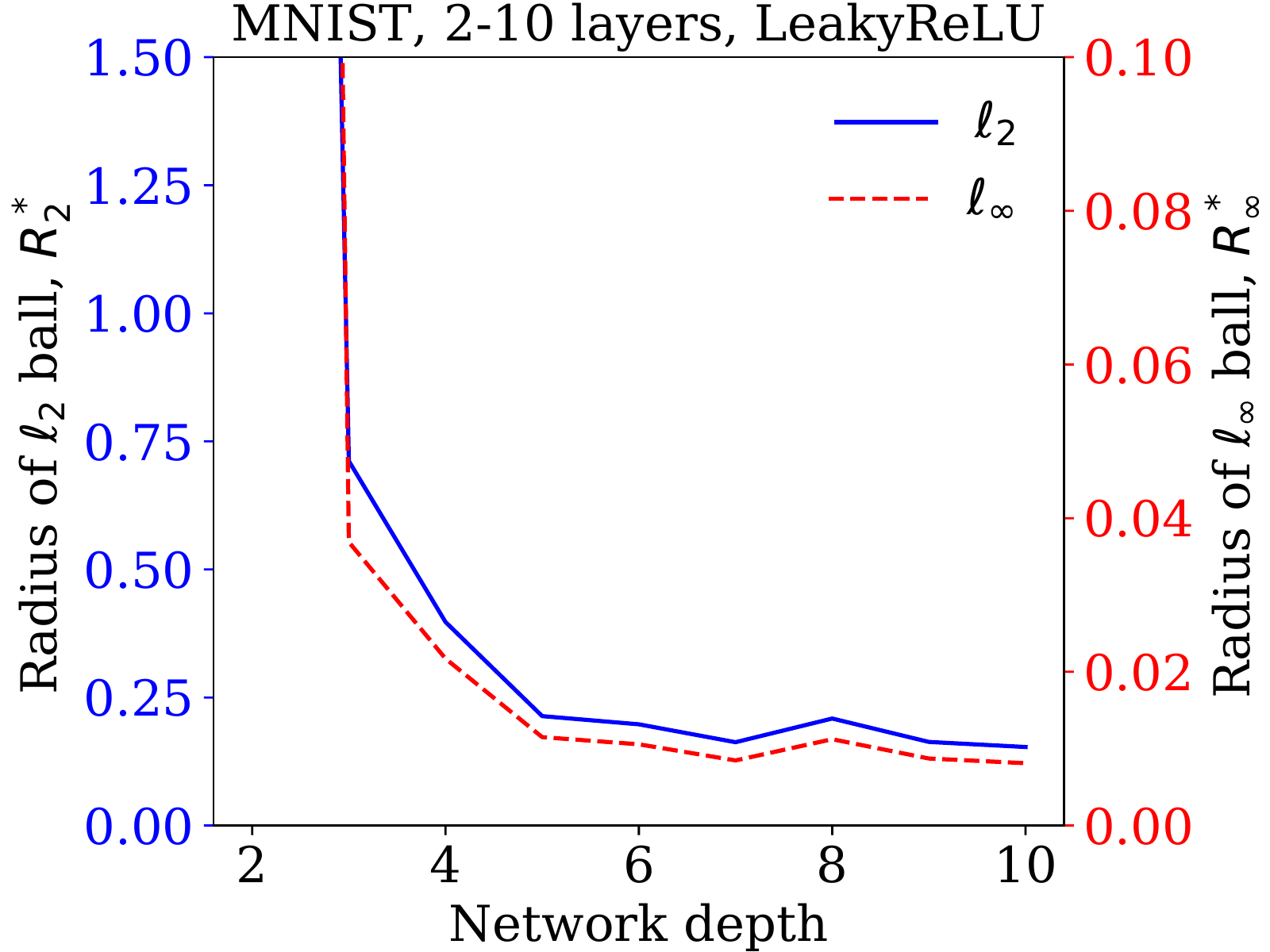}
  \caption{The largest radius $R^*$ within which no stationary point exists, for network with different depths (2-10 layers)}
  \label{fig:landscape}
\end{figure}

In non-convex optimization, a zero gradient vector results in a stationary point, potentially a saddle point or a local minimum. The existence of saddle points and local minima is one of the main difficulties for non-convex optimization~\citep{dauphin2014identifying}, including optimization problems on neural networks. 
However, if for at least one pair of $\{j,k\}$ we have $\gradU{}_{j,k} < 0$ (element $\{j,k\}$ is always negative) or $\gradL{}_{j,k} > 0$ (element $\{j,k\}$ is always positive), the Jacobian $\Y{}$ will never become a zero matrix within a local region.

In this experiment, we train an MLP network with leaky-ReLU activation ($\alpha = 0.3$) for MNIST and varying network depth from 2 to 10. Each hidden layer has 20 neurons, and all models achieve over 96\% accuracy on validation set. We randomly choose 500 images of digit ``1'' from the test set that are correctly classified by all models, and bound the gradient of $f_1(x)$ (logit output for class ``1''). For each image, we record the largest $\ell_2$ and $\ell_\infty$ distortion (denoted as $R^*_2$ and $R^*_\infty$) added such that there is at least one element $k$ in $\nabla f_1(x)$ that can never reach zero (i.e., $\gradU{}_{1,k} < 0$ or $\gradL{}_{1,k} > 0$). The reported $R^*$ are the average of 500 images.

Figure~\ref{fig:landscape} shows how $R^*$ decreases as the network depth increases. Interestingly, for the smallest network with only 2 layers, no stationary point is found in its entire domain ($R^* = \infty$). For deeper networks, the region without stationary points near $x$ becomes smaller, indicating the difficulty of finding optimal adversarial examples (a global optima with minimum distortion) grows with network depth.

\begin{figure*}[ht]
    \centering
    \begin{subfigure}[b]{0.3\textwidth}
        \includegraphics[width=\textwidth]{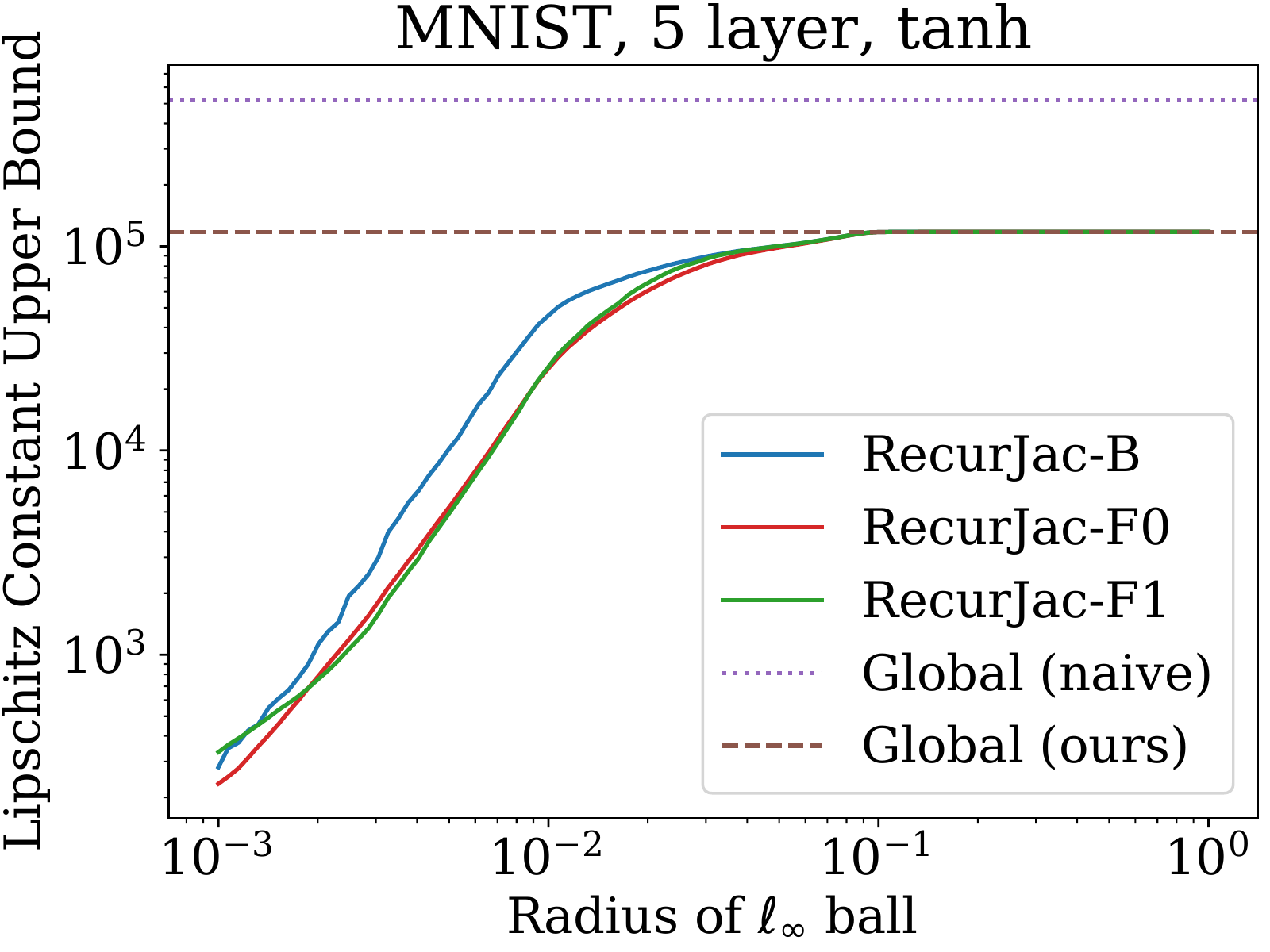}
        \caption{MNIST 5-layer $\tanh$ activation}
        \label{fig:lip_mnist_5_tanh}
    \end{subfigure}
    \begin{subfigure}[b]{0.3\textwidth}
        \includegraphics[width=\textwidth]{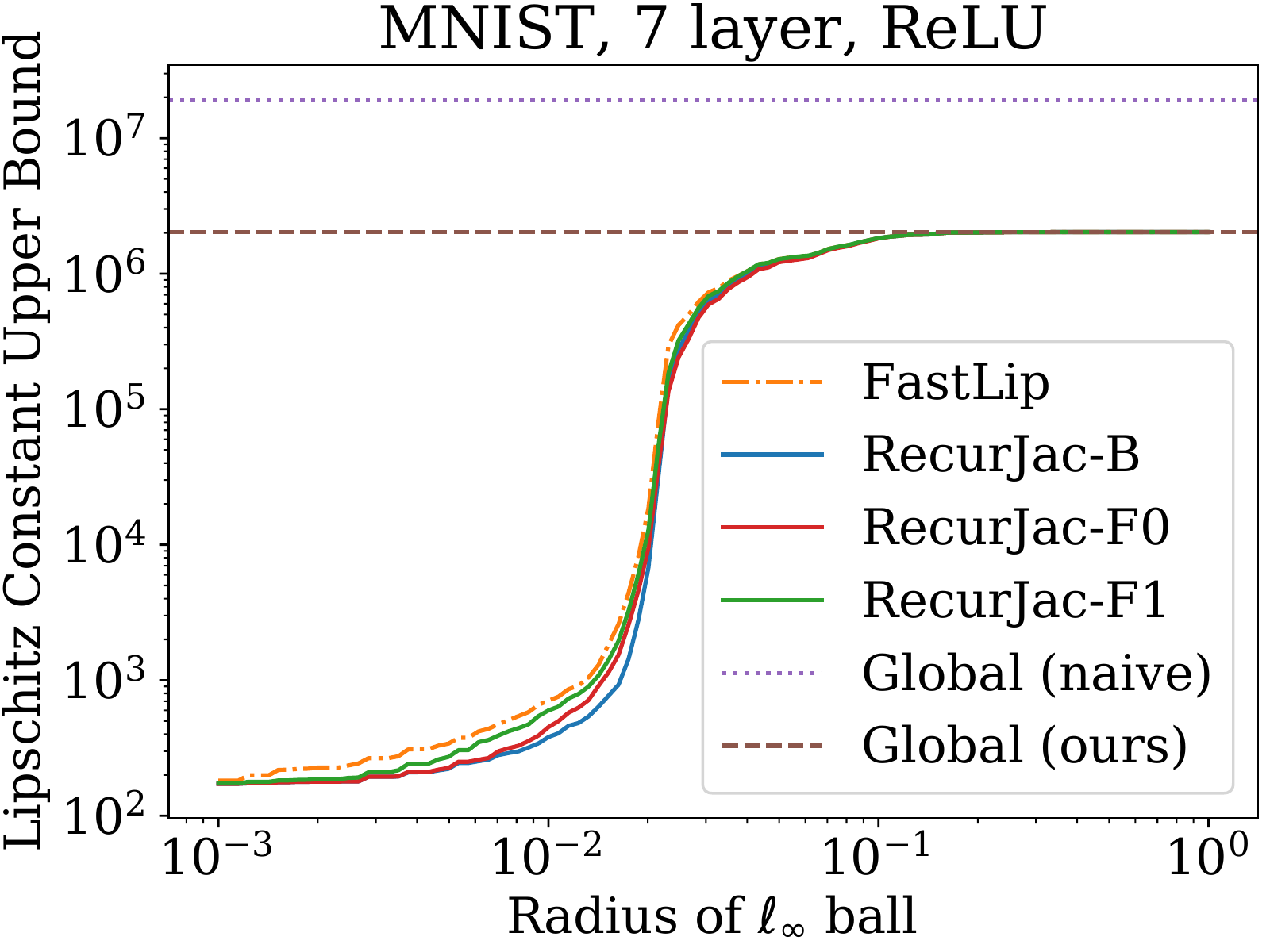}
        \caption{MNIST 7-layer ReLU activation}
        \label{fig:lip_mnist_7_relu}
    \end{subfigure}
    \begin{subfigure}[b]{0.3\textwidth}
        \includegraphics[width=\textwidth]{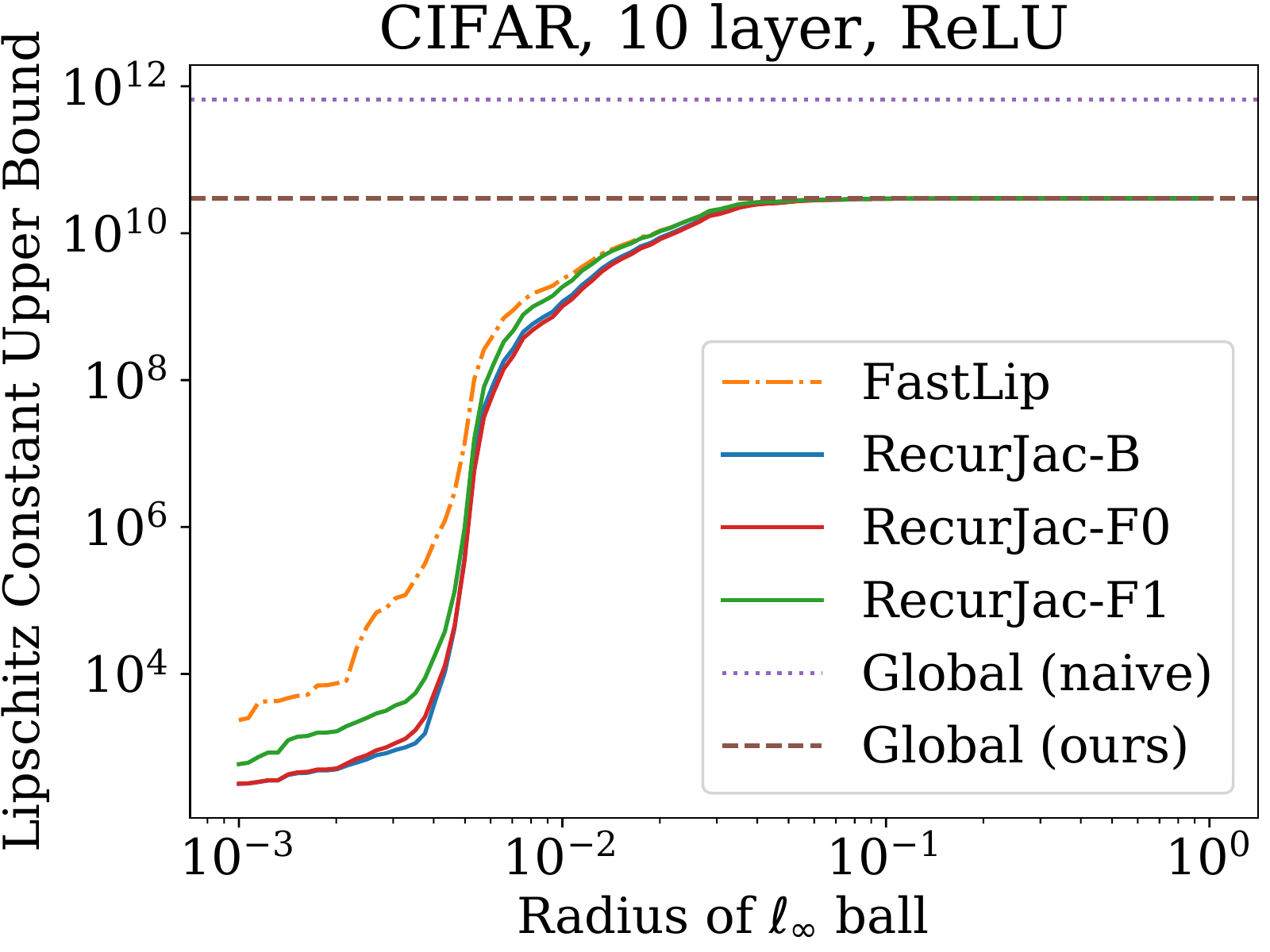}
        \caption{CIFAR 10-layer ReLU activation}
        \label{fig:fig:cifar_10_relu}
    \end{subfigure}
    \caption{Global and local Lipschitz constants on three networks. FastLip can only be applied to (leaky)ReLU networks.}
    \label{fig:lips_large}
\end{figure*}

\begin{table*}[ht]
\centering
% use resize box will automatically resize the table in both single and double column modes
\scalebox{0.8}{
\resizebox{\linewidth}{!}{
\begin{tabular}{|c|c|cc|cc|cc|}
\cline{1-8}
                             & \multicolumn{1}{c|}{}       & \multicolumn{2}{c|}{runner-up target}        & \multicolumn{2}{c|}{random target}           & \multicolumn{2}{c|}{least-likely target}                     \\ \hline
\multicolumn{1}{|c|}{Network} & \multicolumn{1}{c|}{Method} & Undefended &   Adv. Training & Undefended &   Adv. Training & Undefended   & Adv. Training                \\ \hline

\multirow{2}{*}{\makecell{MNIST \\ 3-layer}} & RecurJac & 0.02256 & \bf 0.11573 & 0.02870 & \bf 0.13753 & 0.03205 & \bf 0.16153 \\ 
                                              & FastLip & 0.01802 & 0.09639     & 0.02374 & 0.11753     & 0.02720 & 0.14067     \\ \hline
\multirow{2}{*}{\makecell{MNIST \\ 4-layer}} & RecurJac & 0.02104 & \bf 0.07350 & 0.02399 & \bf 0.08603 & 0.02519 & \bf 0.09863 \\
                                              & FastLip & 0.01602 & 0.04232     & 0.01882 & 0.05267     & 0.02018 & 0.06417     \\ \hline

\end{tabular}}
}
\caption{Comparison of the lower bounds for $\ell_\infty$ distortion found by RecurJac (our algorithm) and FastLip on models with adversarial training with PGD perturbation $\epsilon = 0.3$ for two models and 3 targeted attack classes, averaged over 100 images.}
\label{tb:distill}
\end{table*}

\subsection{Local and global Lipschitz constant}

We apply our algorithm to get local and global Lipschitz constants on four networks of different scales for MNIST and CIFAR datasets. For MNIST, we use a 10-layer leaky-ReLU network with 20 neurons per layer, a 5-layer $\tanh$ network with 50 neurons per layer, a 7-layer ReLU network with 1024, 512, 256, 128, 64 and 32 hidden neurons; for CIFAR, we use a 10-layer network with 2048, 2048, 1024, 1024, 512, 512, 256, 256, 128 hidden neurons.

As a comparison, we include Lipschitz constants computed by Fast-Lip~\citep{weng2018towards}, a state-of-the-art algorithm for ReLU networks (we also trivially extended it to the leaky ReLU case for comparison).
For our algorithm, we run both the backward and the forward versions, denoted as RecurJac-B (Algorithm~\ref{alg:recursive_LU}) and RecurJac-F0 (the forward version). RecurJac-F0 requires to maintain intermediate bounds in shape $n_l \times n_0$, thus the computational cost is very high. We implemented another forward version, RecurJac-F1, which starts intermediate bounds after the first layer and reduce the space complexity to $n_l \times n_1$.

We randomly select an image for each dataset and as the input. Then, we upper bound the Local Lipschitz constant within an $\ell_\infty$ ball of radius $R$. As shown in Figure~\ref{fig:lips_small} and \ref{fig:lips_large}, for all networks, when $R$ is small, our algorithms significantly outperforms Fast-Lip as we find much smaller (and thus in better quality) Lipschitz constants (sometimes a few magnitudes smaller, noting the logarithmic y-axis); When $R$ is large, local Lipschitz constant converges to a value which corresponds to the worst case activation pattern, which is in fact a global Lipschitz constant. Although this value is large, it is still magnitudes smaller than the global Lipschitz constant obtained by the naive product of weight matrices' induced norms (dotted lines with label ``naive''), which is widely used in the neural network literature.

For the largest CIFAR network, the average computation time for 1 local Lipschitz constant of FastLin, RecurJac-B, RecurJac-F0 and RecurJac-F1 are 2.4 seconds, 10.5 seconds, 1 hour and 5 hours respectively, on 1 CPU core. 
RecurJac-F0 and RecurJac-F1 sometimes provide better results than RecurJac-B (Fig.~\ref{fig:lip_mnist_5_tanh}). However in our case when $n_H \ll n_0$, computing the bound in a backward manner is preferred due to its computational efficiency.

\subsection{Robustness verification for adversarial examples}

For a correctly classified source image $s$ of class $c$ and an attack target class $j$, we define $g(s) = f_c(s) - f_j(s) > 0$ that represents the margin between two classes. For $x \in B_{\ell_p}[s; R]$, if $g(x)$ goes below 0, an adversarial example $x$ is found. Using Theorem~\ref{thm:robustness_integral}, we know that the largest $R$ such that $\int_0^R L^{B[s; t]}(g) dt < g(s)$ is a certified robustness lower bound within which no adversarial examples of class $j$ can be found. In this experiment, we approximate the integral in \eqref{eq:approx_lip_integral} from above by using 30 intervals.

We evaluate the robustness lower bound on undefended networks and adversarially trained networks proposed by~\cite{madry2018towards} (which is by far one of the best defending methods). We use two MLP networks with 3 and 4 layers with 1024 neurons per layer.
Table~\ref{tb:distill} shows that our algorithm can indeed reflect the increased robustness as the certified lower bounds under ``Adv. Training'' column  become much larger than ``Undefended''. Additionally, when the adversarial training procedure attempts to defend against adversarial examples with $\ell_\infty$ distortion less than 0.3, our bounds are better than Fast-Lip and closer to 0.3, suggesting that adversarial training is an effective defense.

\section{Conclusion}

In this paper, we propose a novel algorithm, RecurJac, for recursively bounding a neural network's Jacobian matrix with respect to its input. Our method can be efficiently applied to networks with a wide class of activation functions. We also demonstrate several applications of our bounds in experiments, including characterizing local optimization landscape, computing a local or global Lipschitz constant, and robustness verification of neural networks.

\clearpage

\paragraph{Acknowledgment.}

The authors thank Zeyuan Allen-Zhu, Sebastien Bubeck, Po-Sen Huang, Kenji Kawaguchi, Jason D. Lee, Suhua Lei, Xiangru Lian, Mark Sellke, Zhao Song and Tsui-Wei Weng for discussing ideas in this paper and providing insightful feedback. The authors also acknowledge the support of NSF via IIS-1719097, Intel, Google Cloud and Nvidia.

\bibliography{all}
\bibliographystyle{arxiv}

\clearpage

\appendix
\begin{center}
\textbf{\LARGE Appendix}
\end{center}
\section{Proofs}
\paragraph{Proof of Proposition~\ref{prop:recursion_I}}
\begin{proof}
In \eqref{eq:multilayer_decom}, we define $\setYo{j}:=\{r ~:~ \gradL{(-l)}_{j,r} <0 < \gradU{(-l)}_{j,r}\}$. Then we can decompose $I$ into two terms:
\begin{align*}
\begin{split}
I = & \sum_{r\in \setYo{j}, \W{(l-1)}_{r,k}<0} \Y{(-l)}_{j,r}\sigma'(f_r^{(l-1)}(\x)) \W{(l-1)}_{r,k} %\\ & 
+ \sum_{r\in \setYo{j}, \W{(l-1)}_{r,k}>0} \Y{(-l)}_{j,r}\sigma'(f_r^{(l-1)}(\x)) \W{(l-1)}_{r,k}
\end{split}
\end{align*}
Since $\sigma'(f_r^{(l-1)}(\x))$ is always non-negative, a lower bound of the first term is $\gradU{(-l)}_{j,r} \gradupbnd{(l-1)}_r \W{(l-1)}_{r,k}$, and a lower bound of the second term is $\gradL{(-l)}_{j,r} \gradupbnd{(l-1)}_r \W{(l-1)}_{r,k}$. Notice that in both terms we take $\gradupbnd{(l-1)}_r$ because both $\gradU{(-l)}_{j,r} \W{(l-1)}_{r,k}  (\text{when } \W{(l-1)}_{r,k}<0)$ and $\gradL{(-l)}_{j,r} \W{(l-1)}_{r,k} (\text{when } \W{(l-1)}_{r,k}>0)$ are non-positive when $r\in \setYo{j}$. Therefore, $\gradL{(-l+1),\pm}_{j,k}$ defined in \eqref{eq:nlayer_lb_I} is a lower bound for term $I$.

Similarly, we can show that $\gradU{(-l+1),\pm}_{j,k}$ defined in \eqref{eq:nlayer_ub_I} is an upper bound for term $I$.
\end{proof}

\paragraph{Proof of Lemma~\ref{lem:recursion_II}}
\begin{proof}
By definition in \eqref{eq:multilayer_decom}, we have
\begin{align*}
\begin{split}
II &= \sum_{\{r ~:~ \gradL{(-l)}_{j,r} \ge 0 \text{ or } \gradU{(-l)}_{j,r} \le 0\}} \Y{(-l)}_{j,r}\sigma'(f_r^{(l-1)}(\x)) \W{(l-1)}_{r,k} \\
& = \sum_{\stackrel{\gradL{(-l)}_{j,r} \ge 0, \W{(l-1)}_{r,k}>0}{\text{or } \gradU{(-l)}_{j,r} \le 0, \W{(l-1)}_{r,k}<0}} \Y{(-l)}_{j,r}\sigma'(f_r^{(l-1)}(\x)) \W{(l-1)}_{r,k} %\\ & 
+ \sum_{\stackrel{\gradL{(-l)}_{j,r} \ge 0, \W{(l-1)}_{r,k}<0}{\text{or } \gradU{(-l)}_{j,r} \le 0, \W{(l-1)}_{r,k}>0}} \Y{(-l)}_{j,r}\sigma'(f_r^{(l-1)}(\x)) \W{(l-1)}_{r,k}.
\end{split}
\end{align*}
In the first term $\Y{(-l)}_{j,r} \W{(l-1)}_{r,k}$ is sure to be non-negative, while in the second term it is sure to be non-positive. Since $\sigma'(f_r^{(l-1)}(\x))$ is always non-negative, we can obtain a lower bound as below:
\begin{align*}
\begin{split}
& II \ge \sum_{\stackrel{\gradL{(-l)}_{j,r} \ge 0, \W{(l-1)}_{r,k}>0}{\text{or } \gradU{(-l)}_{j,r} \le 0, \W{(l-1)}_{r,k}<0}} \Y{(-l)}_{j,r} \gradlwbnd{(l-1)}_r \W{(l-1)}_{r,k}
% \\ &
+ \sum_{\stackrel{\gradL{(-l)}_{j,r} \ge 0, \W{(l-1)}_{r,k}<0}{\text{or } \gradU{(-l)}_{j,r} \le 0, \W{(l-1)}_{r,k}>0}} \Y{(-l)}_{j,r} \gradupbnd{(l-1)}_r \W{(l-1)}_{r,k}.
\end{split}
\end{align*}
Plugging $\Y{(-l)}_{j,r} = \sum_{s} \Y{(-l-1)}_{j,i} \sigma'(f_i^{(l)}(\x)) \W{(l)}_{i,r}$ into the lower bound above and changing the summation order between $i$ and $r$, we obtain
\begin{align*}
\begin{split}
& II \ge \sum_{i} \hspace{-2mm} \sum_{\stackrel{\gradL{(-l)}_{j,r} \ge 0, \W{(l-1)}_{r,k}>0}{\text{or } \gradU{(-l)}_{j,r} \le 0, \W{(l-1)}_{r,k}<0}} \hspace{-4mm} \Y{(-l-1)}_{j,i} \sigma'(f_i^{(l)}(\x)) \W{(l)}_{i,r} \gradlwbnd{(l-1)}_r \W{(l-1)}_{r,k} \\
& + \sum_i \hspace{-2mm} \sum_{\stackrel{\gradL{(-l)}_{j,r} \ge 0, \W{(l-1)}_{r,k}<0}{\text{or } \gradU{(-l)}_{j,r} \le 0, \W{(l-1)}_{r,k}>0}} \hspace{-4mm} \Y{(-l-1)}_{j,i} \sigma'(f_i^{(l)}(\x)) \W{(l)}_{i,r} \gradupbnd{(l-1)}_r \W{(l-1)}_{r,k}.
\end{split}
\end{align*}
Note that $j,k$ are fixed numbers and do not change in the summation, and the inner summation is only over $r$. After merging terms, the lower bound above is exactly $\Y{(-l-1)}_{j,:} \Sigma^{(l)} \checkW{(l,l-1,j)}_{:,k}$ where $\checkW{(l,l-1,j)}_{:,k}$ is defined in \eqref{eq:multilayer_lb_II}. 

Similarly, we can show that $\Y{(-l-1)}_{j,:} \Sigma^{(l)} \hatW{(l,l-1,j)}_{:,k}$ is an upper bound for term $II$, where $\hatW{(l,l-1,j)}_{:,k}$ is defined in \eqref{eq:multilayer_ub_II}.
\end{proof}

\paragraph{Proof of Theorem~\ref{thm:multiple_LU}}
\begin{proof}
Combining Equation~\eqref{eq:multilayer}, Proposition~\ref{prop:recursion_I} and Lemma~\ref{lem:recursion_II}, the results in Theorem~\ref{thm:multiple_LU} immediately follow. 
\end{proof}

\paragraph{Proof of Proposition~\ref{prop:Lp_bnd}}
\begin{proof}
For $1 \le p < +\infty$, by the definition of induced norm: 
\begin{align*}
\begin{split}
\|\nabla f(x)\|_p &= \sup_{\|y\|_p = 1} \| \Y{} y \|_p = \sup_{\|y\|_p = 1} \left( \sum_{j} \left( \left | \sum_{k} \Y{}_{j, k} y_k \right| \right)^p \right)^{\frac{1}{p}} \\
& \le \sup_{\|y\|_p = 1} \left( \sum_{j} \left( \sum_{k} \left |\Y{}_{j, k}\right| |y_k| \right)^p \right)^{\frac{1}{p}} \\
& \le \sup_{\|y\|_p = 1} \left( \sum_{j} \left( \sum_{k} \M_{j,k} |y_k| \right)^p \right)^{\frac{1}{p}} \\
& \le \| M \|_{p}.
\end{split}
\end{align*}
Since the inequality holds for any $x \in B_{\ell_p}[s; R]$, we obtain \eqref{eq:Lp_bnd} for $1\le p < +\infty$.

For $p=\infty$, by the definition of induced norm: 
\begin{align*}
\begin{split}
\|\nabla f(x)\|_{\infty} &= \max_j \sum_{k} | \Y{}_{j,k} | \le \max_j \sum_{k} | \M_{j,k} | = \| M \|_{\infty}.
\end{split}
\end{align*}
Since the inequality holds for any $ x \in B_{\ell_p}[s;R] $, we obtain \eqref{eq:Lp_bnd} for $p = +\infty$.
\end{proof}

\paragraph{Proof of Theorem~\ref{thm:robustness_integral}}
\begin{proof}
First consider all points $x$ where $\| x - s \| = R$.
We define $x_0 = s$, $x_n = x$, and a function $\bm{r}(t) = x_0 + \frac{t}{R} (x_n - x_0)$. Splitting the line between $x_0$ and $x_n$ by $n$ pieces, where $t_i = \frac{i}{n} R$, $\| x_i - x_0 \| = t_i$, we have

\begin{align}
\| f(x_n) - f(x_0) \| & \leq \sum_{i=1}^{n} \| f(x_i) - f(x_{i-1}) \| \nonumber \\
& \leq \sum_{i=1}^{n} L^{B[s; \frac{i}{n} R]} \| x_i - x_{i-1} \| \nonumber \\
& \leq \sum_{i=1}^{n} L^{B[s; t_i]} \| \bm{r}(t_i) - \bm{r}(t_{i-1}) \| \nonumber \\
\label{eq:lips_integral_finite}
&= \sum_{i=1}^{n} L^{B[s; t_i]} \| \bm{r'}(t_i) \| \Delta t
\end{align}

\eqref{eq:lips_integral_finite} is equality due to the linearity of $\bm{r}(t_i)$. Noticing that $\| \bm{r'}(t_i) \| = \| \frac{x_n - x_0}{R} \| = 1$, when $n \rightarrow \infty$, $ \Delta t \rightarrow 0$, \eqref{eq:lips_integral_finite} becomes a Riemann integral:

\[
\lim_{n \rightarrow \infty} \sum_{i=1}^{n} L^{B[s; t_i]} \| \bm{r'}(t_i) \| \Delta t = \int_0^R L^{B[s; t]} dt
\]

For any $x'$ where $\| x' - s \| = R_1 \leq R$, due to the non-negativity of Lipschitz constant, we have 

\[
\| f(x') - f(s) \| \leq \int_0^{R_1} L^{B[s; t]} dt \leq \int_0^R L^{B[s; t]} dt \leq L^{B[s; R]} \cdot R
\]

The last ``$\leq$'' sign holds as Lipschitz constant is non-decreasing when $t$ increases.

\end{proof}

\end{document}